\documentclass{article}

\usepackage[final]{neurips_2023}

\usepackage[utf8]{inputenc} % allow utf-8 input
\usepackage[T1]{fontenc}    % use 8-bit T1 fonts
\usepackage{hyperref}       % hyperlinks
\usepackage{url}            % simple URL typesetting
\usepackage{booktabs}       % professional-quality tables
\usepackage{amsfonts}       % blackboard math symbols
\usepackage{nicefrac}       % compact symbols for 1/2, etc.
\usepackage{microtype}      % microtypography
\usepackage{xcolor}         % colors
\usepackage{graphicx}
\usepackage{subfigure}
\usepackage{makecell}
\usepackage{enumitem}

\title{Efficient LLM Inference on CPUs}

\author{%
Haihao Shen \quad Hanwen Chang \quad Bo Dong \quad Yu Luo \quad Hengyu Meng \\
\texttt{\{haihao.shen, hanwen.chang, bo1.dong, yu.luo, hengyu.meng\}@intel.com}
}

\begin{document}

\maketitle

\begin{abstract}
Large language models (LLMs) have demonstrated remarkable performance and tremendous potential across a wide range of tasks. However, deploying these models has been challenging due to the astronomical amount of model parameters, which requires a demand for large memory capacity and high memory bandwidth. In this paper, we propose an effective approach that can make the deployment of LLMs more efficiently. We support an automatic INT4 weight-only quantization flow and design a special LLM runtime with highly-optimized kernels to accelerate the LLM inference on CPUs. We demonstrate the general applicability of our approach on popular LLMs including Llama2, Llama, GPT-NeoX, and showcase the extreme inference efficiency on CPUs. The code is publicly available at: https://github.com/intel/intel-extension-for-transformers.
\end{abstract}

\section{Introduction}\label{intro}
Large language models (LLMs) have shown remarkable performance and tremendous potential across a wide range of tasks~\citet{roziere2023code, touvron2023llama2, touvron2023llama, zhang2022opt, brown2020language, li2023starcoder}. However, deploying these models has been challenging due to the astronomical amount of model parameters, which necessitates significant memory capacity and high memory bandwidth.

Quantization is a technique to reduce the numeric precision of weights and activations of a neural network to lower the computation costs of inference. INT8 quantization~\citet{int8_vanhoucke, songhan_deep_compression, Jacob_2018_CVPR} is the most widely-used approach today given the trade-off between high inference performance and reasonable model accuracy. However, outliers in activations have been observed and those outlier values are limiting the general adoption of INT8 quantization, though there are some related work that has been proposed to address the issues~\citet{xiao2023smoothquant, wei2023outlier, dettmers2022llm}. FP8 is a newly introduced data type that has attracted lots of attentions~\citet{micikevicius2022fp8, kuzmin2022fp8, sun2019hybrid, shen2023efficient} while it has little adoptions due to the hardware unavailability. On the other hand, weight-only quantization becomes popular as it applies the low precision (e.g., 4-bit) to weights only, while keeping higher precision (e.g., 16-bit floating point) for activations, therefore maintaining the model accuracy. There are many excellent work on 4-bit weight-only quantization~\citet{dettmers2023qlora, frantar2022gptq, cheng2023optimize, lin2023awq, kim2023squeezellm, wu2023zeroquant, cheng2023teq} that have demonstrated the effectiveness in LLM inference. Meanwhile, the open-source community is embracing such low-bit weight-only quantization and offers the CPP-based implementations such as~\href{https://github.com/ggerganov/llama.cpp}{llama.cpp} and~\href{https://github.com/bigcode-project/starcoder.cpp}{starcoder.cpp} based on ~\href{https://github.com/ggerganov/ggml}{ggml} library. These implementations are typically optimized for CUDA and may not work on CPUs. Therefore, it is important to address the challenge of making LLM inference efficient on CPU.

In this paper, we propose an effective approach for LLM inference on CPUs including an automatic INT4 quantization flow and an efficient LLM runtime. We leverage~\href{https://github.com/intel/neural-compressor}{Intel Neural Compressor} that provides the support of INT4 quantization such as GPTQ~\citet{frantar2022gptq}, AWQ~\citet{lin2023awq}, TEQ~\citet{cheng2023teq}, SignRound~\citet{cheng2023optimize} and generate the INT4 model automatically. Inspired from the~\href{https://github.com/ggerganov/ggml}{ggml} library, we develop a tensor library for CPU, supporting all the mainstream instruction sets such as AVX2, AVX512, AVX512\_VNNI~\citet{rodriguez2018lower}, and AMX (\href{https://www.intel.com/content/www/us/en/products/docs/accelerator-engines/advanced-matrix-extensions/overview.html}{Advanced Matrix Extensions}). Our results show the average latency of generation tokens from 20ms to 80ms on LLMs with 6B to 20B parameters using just a single socket of 4th Generation Intel® Xeon® Scalable Processors, while preserving the high accuracy within only 1\% loss from FP32 baseline. Our main contributions are as follows:

\begin{itemize}
    \item We propose an automatic INT4 quantization flow and generate the high-quality INT4 models with negligible accuracy loss within <1\% from FP32 baseline.
    \item We design a tensor library that supports general CPU instruction sets and latest instruction sets for deep learning acceleration. With CPU tensor library, we develop an efficient LLM runtime to accelerate the inference.
    \item We apply our inference solution to popular LLM models covering 3B to 20B and demonstrate the promising per-token generation latency from 20ms to 80ms, much faster than the average human reading speed~\href{https://www.quora.com/What-is-the-reading-speed-of-an-average-person}{about 200ms per token}.
\end{itemize}

The rest of this paper is organized as follows. Section~\ref{sec:approach} introduces the approach including INT4 quantization and inference. Section~\ref{sec:results} outlines the experimental setup, presents accuracy \& performance results, and offers discussion on performance tuning. Section \ref{sec:summary} presents the conclusions and future work.

\section{Approach}
\label{sec:approach}
\begin{figure*}[htbp]
\centering
\includegraphics[width=1.0\linewidth]{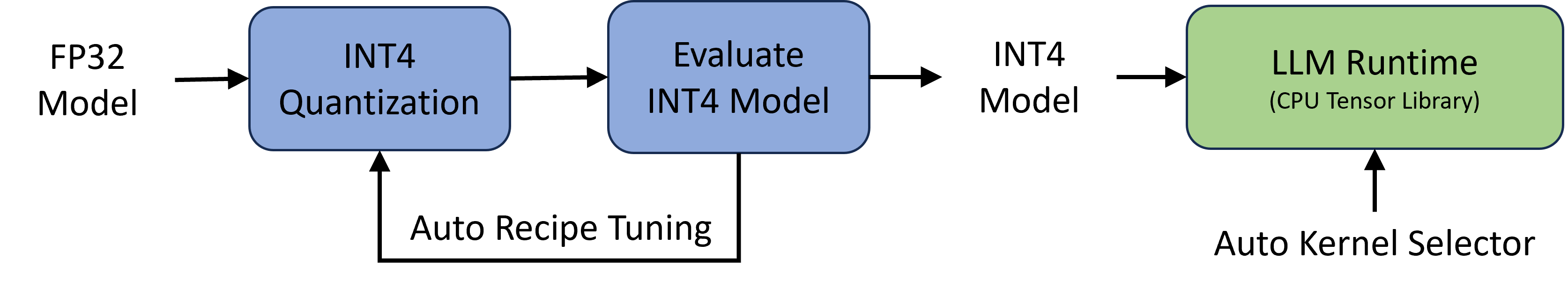}
\caption{The left part is the automatic INT4 quantization flow: given a FP32 model, the flow takes the default INT4 quantization recipes and evaluates the accuracy of INT4 model; the recipe tuning loop is optional, if INT4 model can meet the accuracy target. The right part is a simplified runtime for efficient LLM inference built on top of a CPU tensor library with automatic kernel selector.}
\label{fig:architecture}
\end{figure*}

In this section, we introduce the approach which consists of two major components: an automatic INT4 quantization flow and an efficient LLM runtime, as shown in Figure~\ref{fig:architecture}. More details are described in the following sections.

\subsection{Automatic INT4 Quantization Flow}
\label{int4_quantization}
INT4 quantization flow is developed based on Intel Neural Compressor, a popular quantization tool for deep learning frameworks. Since the tool has already supported the mainstream INT4 quantization recipes such as GPTQ, SignRound, AWQ, TEQ, and RTN (round-to-nearest), our automatic quantization flow allows the recipe tuning on different quantization recipes, different granularities (channel-wise or group-wise), different group size (32, 64, 128 ... 1024). Each recipe generates an INT4 model that is evaluated in the flow. Once the INT4 model meets the accuracy target, the model will be passed to LLM Runtime for performance evaluation.

\subsection{Efficient LLM Runtime}
\label{llm_runtime}
\begin{figure*}[htbp]
\centering
\includegraphics[width=1.0\linewidth]{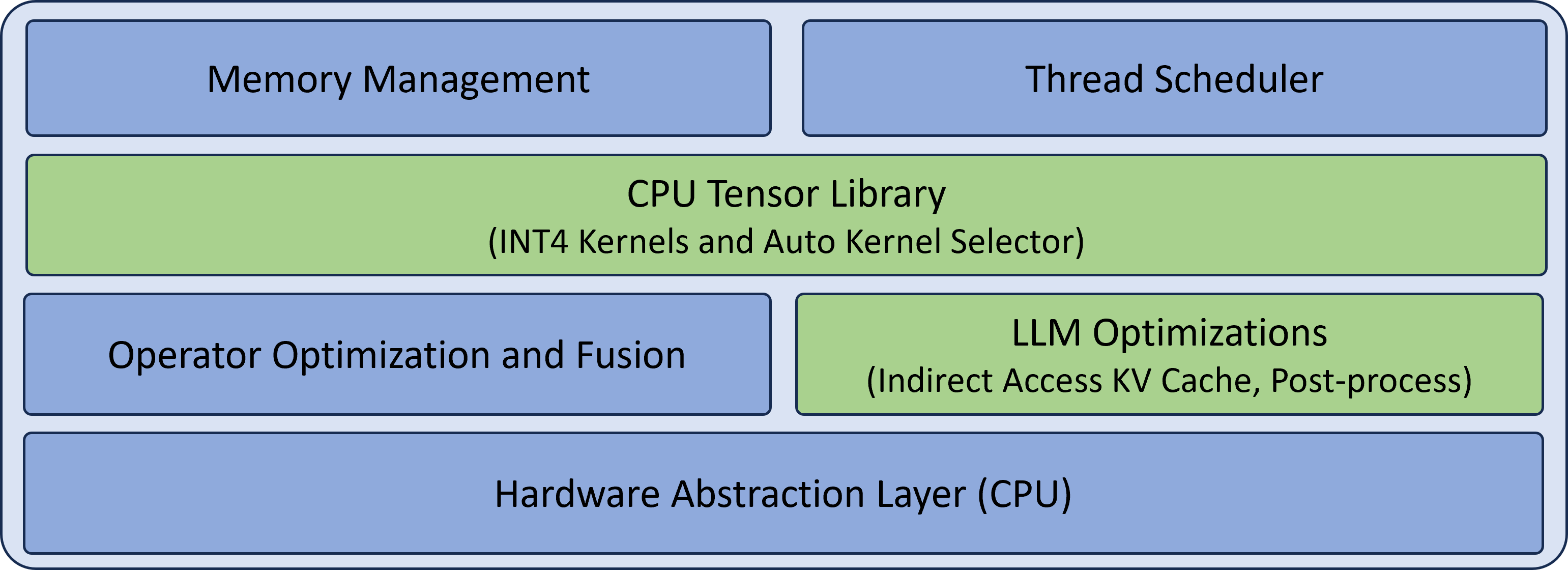}
\caption{Key components in LLM runtime: general and LLM specialized.}
\label{fig:llm_runtime}
\end{figure*}

LLM runtime is designed to provide the efficient inference of LLMs on CPUs. Figure~\ref{fig:llm_runtime} describes the key components in LLM runtime, where the components (CPU tensor library and LLM optimizations) in green are specialized for LLM inference, while the other components (memory management, thread scheduler, operator optimization and fusion) in blue are required for a general runtime. More details about CPU tensor library and LLM optimizations are described in the following paragraphs, while the general components are omitted due to the space limitations. Note that the design is flexibly extensible with hardware abstraction layer (CPU only for now), while how to support other hardware is out of scope in this paper. 

\paragraph{CPU Tensor Library.} We develop CPU tensor library for linear algebra subroutines, inspired from the template design of~\href{https://github.com/NVIDIA/cutlass}{cutlass}. The tensor library offers a comprehensive support of INT4 kernels for x86 CPUs as shown in Table~\ref{tbl:tensor_library}, where AMX is available in the latest Intel Xeon Scalable Processors and VNNI is available in both Intel and AMD CPUs.

\begin{table}[htbp]
\caption{Support matrix by CPU tensor library: input/output data type, compute data type, and ISA (instruction set architecture). The library supports dynamic quantization for input along with batch or input channel per group, and  weight quantization in both symmetric and asymmetric scheme.}
\label{tbl:tensor_library}
\centering
\begin{tabular}{|c|c|c|c|}
\hline
Input Data Type  &  Output Data Type & Compute Data Type  & Compute ISA \\ \hline
FP32  & FP32 & FP32 & AVX2 \\ \hline
FP32  & FP32 & FP32 & AVX512F \\ \hline
FP32  & FP32 & INT8 & AVX\_VNNI \\ \hline
FP32  & FP32 & INT8 & AVX512\_VNNI \\ \hline
FP32  & FP32 & INT8 & AMX\_INT8 \\ \hline
FP32/FP16  & FP32/FP16 & FP16 & AVX512\_FP16 \\ \hline
FP32/BF16  & FP32/BF16 & BF16 & AMX\_BF16 \\ \hline
\end{tabular}
\end{table}

\begin{figure}[htbp]
	\begin{minipage}[c]{0.45\linewidth}
		\vspace{5pt}
		\centerline{\includegraphics[width=\textwidth]{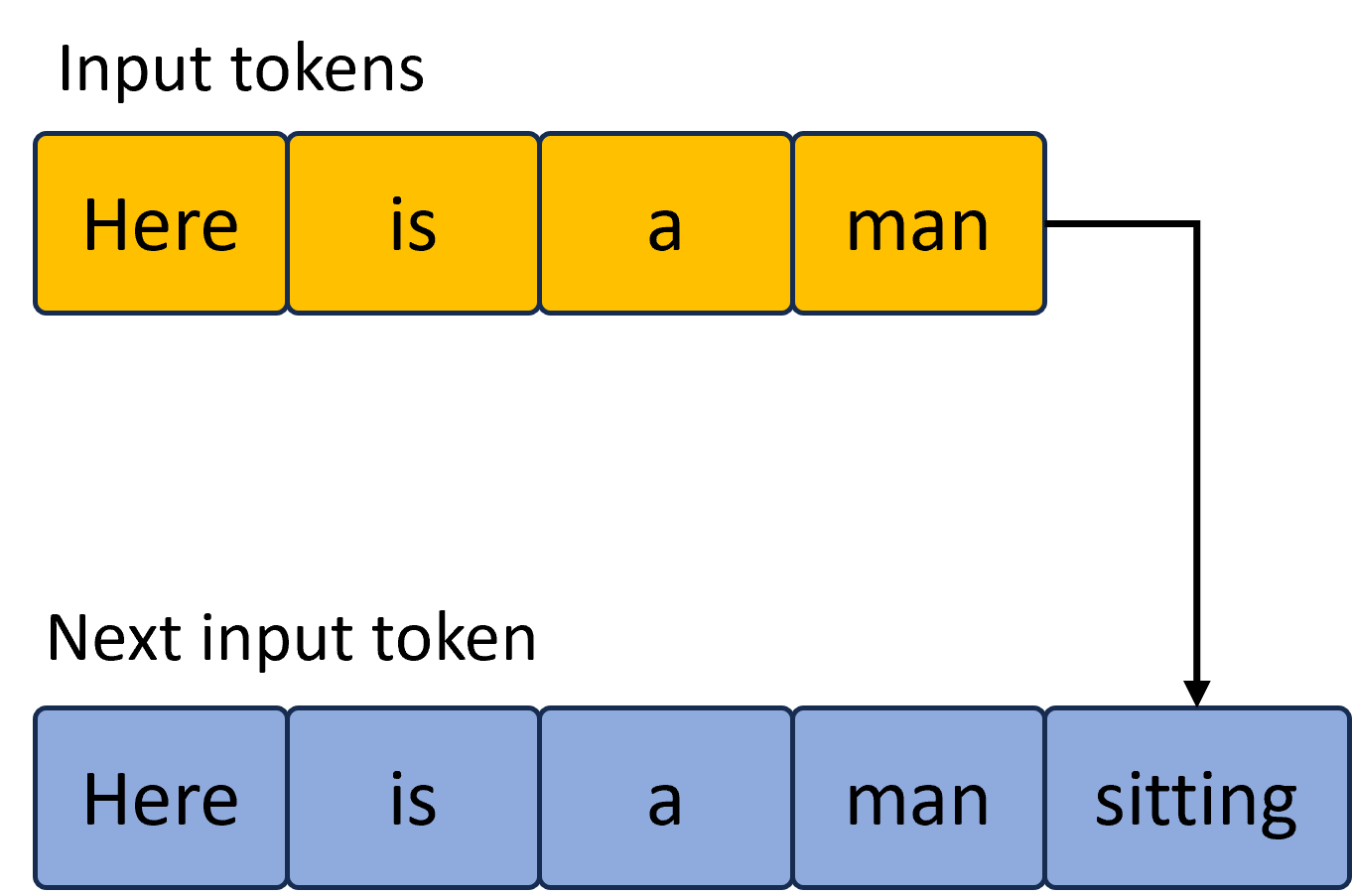}}
		 \centerline{(a)}
	\end{minipage}
	\begin{minipage}[c]{0.5\linewidth}
		\vspace{5pt}
		\centerline{\includegraphics[width=\textwidth]{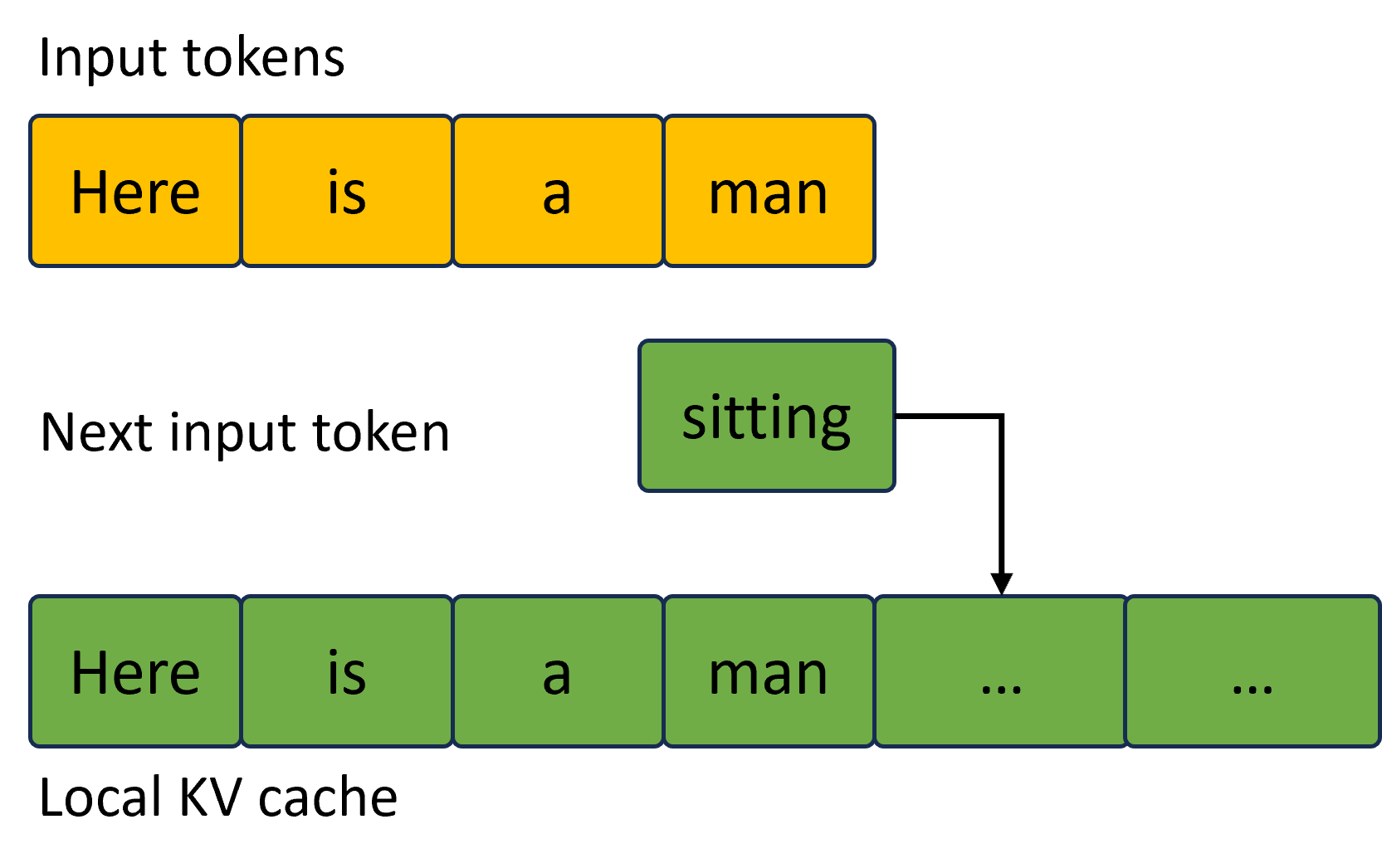}}
		\centerline{(b)}
	\end{minipage}
	\caption{KV cache optimization. Left (a) shows the default KV cache, where new token generation requires memory reallocation for all the tokens (5 in this example); right (b) shows the optimized KV cache with pre-allocated KV memory and only new token updated each time.}
\label{fig:kv_cache}
\end{figure}

\paragraph{LLM Optimizations.} Most recent LLMs are typically decoder-only Transformer-based models~\citet{vaswani2017attention}. Given the unique characteristics of next token generation, KV cache becomes performance critical for LLM inference. We describe the optimizations in Figure~\ref{fig:kv_cache}.

\section{Results}
\label{sec:results}
%We describe the experimental setup and present the accuracy and performance results as follows.

\subsection{Experimental Setup}
\label{sec:setup}
To demonstrate the generality, we select the popular LLMs across a wide range of architectures with the model parameter size from 7B to 20B. We evaluate the accuracy of both FP32 and INT4 models using open-source datasets from \href{https://github.com/EleutherAI/lm-evaluation-harness}{lm-evaluation-harness} including lambada~\citet{paperno2016lambada} openai, hellaswag~\citet{zellers2019hellaswag}, winogrande~\citet{sakaguchi2021winogrande}, piqa~\citet{bisk2020piqa}, and \href{https://huggingface.co/datasets/wikitext}{wikitext}. To demonstrate the performance, we measure the latency of next token generation on the 4th Generation Intel® Xeon® Scalable Processors, available on the public clouds such as \href{https://aws.amazon.com/}{AWS}.

\subsection{Accuracy}
\label{sec:accuracy}
We evaluate the accuracy on the aforementioned datasets and show the average accuracy in Table~\ref{accuracy}. We can see from the table that the accuracy of INT4 model is nearly on par with that of FP32 model within 1\% relative loss from FP32 baseline.

\begin{table}[htbp]
\caption{INT4 and FP32 model accuracy. INT4 model has two configurations: group size=32 and 128.}
\label{accuracy}
\centering
\begin{tabular}{|c|c|c|c|}
\hline
LLM & FP32   & INT4 (Group size=32) & INT4 (Group size=128) \\ \hline
EleutherAI/gpt-j-6B & 0.643 & 0.644 & 0.64 \\ \hline
meta-llama/Llama-2-7b-hf & 0.69 & 0.69 & 0.685            \\ \hline
decapoda-research/llama-7b-hf  & 0.689 & 0.682 & 0.68     \\ \hline
EleutherAI/gpt-neox-20b  & 0.674 & 0.672 & 0.669          \\ \hline
%mosaicml/mpt-7b   & 0.689 & 0.688 & 0.685          \\ \hline
tiiuae/falcon-7b   & 0.698 & 0.694 & 0.693          \\ \hline
%baichuan-inc/baichuan-7B   & 0.474 & 0.471 & 0.47          \\ \hline
%facebook/opt-6.7b   & 0.65 & 0.647 & 0.643          \\ \hline
%databricks/dolly-v2-3b    & 0.613 & 0.609 & 0.609          \\ \hline
%tiiuae/falcon-40b-instruct    & 0.756 & 0.757 & 0.755     \\ \hline
\end{tabular}
\vspace{-3mm}
\end{table}

\subsection{Performance}
\label{sec:performance}
We measure the latency of next token generation using LLM runtime and the popular open-source ggml-based implementation. Table~\ref{tbl:perf} presents the latency under a proxy configuration with 32 as both input and output tokens. Note that ggml-based solution only supports group size 32 when testing.

\begin{table}[ht]
\caption{INT4 performance using LLM runtime and ggml-based solution. LLM runtime outperforms ggml-based solution by up to 1.6x under group-size=128 and 1.3x under group size=32.}
\label{tbl:perf}
\centering
\begin{tabular}{|c|c|c|c|}
\hline
model & \begin{tabular}[c]{@{}c@{}}LLM Runtime\\ (Group size=32)\end{tabular} & \begin{tabular}[c]{@{}c@{}}LLM Runtime\\ (Group size=128)\end{tabular} & \begin{tabular}[c]{@{}c@{}}ggml-based \\ (Group size=32)\end{tabular} \\ \hline
EleutherAI/gpt-j-6B  & 22.99ms                                                               & 19.98ms                                                                      & 31.62ms                                                                             \\ \hline
meta-llama/Llama-2-7b-hf  & 23.4ms                                                               & 21.96ms                                                                  & 27.71ms                                                                             \\ \hline
decapoda-research/llama-7b-hf  & 23.88ms                                                               & 22.04ms                                                                      & 27.2ms                                                                             \\ \hline
EleutherAI/gpt-neox-20b  & 80.16ms                                                               & 61.21ms                                                                      & 92.36ms                                                                             \\ \hline
%mosaicml/mpt-7b & 25.7ms                                                               & 21.05ms                                                                      & 31.54ms                                                                             \\ \hline

tiiuae/falcon-7b  & 31.23ms                                                               & 22.26ms                                                                      & 36.22ms                                                                             \\ \hline
\end{tabular}
\end{table}

\subsection{Discussion}
\label{sec:discussion}
Though we demonstrate the performance advantage over ggml-based solution, there are still opportunities for LLM runtime to further improve the performance through additional performance tuning such as thread scheduler in LLM runtime, blocking strategy in CPU tensor library.

\section{Summary and Future Work}
\label{sec:summary}
We presented an end-to-end INT4 LLM inference including an automatic INT4 model quantization and efficient LLM runtime. We demonstrated the generality on a set of popular LLMs and the performance advantage over the open-source solution on CPUs. As our future works, we plan to further improve the CPU tensor library and extend Hugging Face transformer APIs to support INT4 LLM inference as part of the contributions to the open-source community. Moreover, we plan to exercise our approach on personal computers (PCs) given the broad accessibility of CPUs, to meet the growing demands of AI generated content and empower generative AI on PCs.

\subsection{Appendix}
\label{sec:append}
Memory usage is a critical metric for the 4-bit solution. The results are presented below.

\begin{table}[htbp]
\caption{INT4 and FP32 memory usage. INT4 model has two configurations: group size=128 and 32.}
\label{memory}
\centering
\begin{tabular}{|c|c|c|c|}
\hline
LLM & FP32   & INT4 (Group size=128) & INT4 (Group size=32) \\ \hline
EleutherAI/gpt-j-6B & 22481MB & 3399MB & 4017MB \\ \hline
meta-llama/Llama-2-7b-hf & 22057MB & 3772MB & 4928MB            \\ \hline
decapoda-research/llama-7b-hf  & 21750MB & 3773MB & 4874MB     \\ \hline
EleutherAI/gpt-neox-20b  & 68829MB & 11221MB & 13458MB          \\ \hline
tiiuae/falcon-7b   & 24624MB & 5335MB & 5610MB          \\ \hline
\end{tabular}
\vspace{-3mm}
\end{table}

\bibliography{paper}
\bibliographystyle{abbrvnat}

\end{document}